\documentclass[a4paper]{article}

\usepackage{nnsp2e}
\usepackage{graphics}

\newcommand{\A}{{\bf A}}
\newcommand{\K}{{\bf K}}
\newcommand{\ai}{{\bf a}_i}

\newcommand{\x}{{\bf x}}
\newcommand{\s}{{\bf s}}
\newcommand{\cvec}{{\bf c}}
\newcommand{\st}{{\bf s}^t}
\newcommand{\X}{{\bf X}}
\newcommand{\bS}{{\bf S}}
\newcommand{\Aorig}{{\bf A}_{\mbox{\footnotesize orig}}}
\newcommand{\Sorig}{{\bf S}_{\mbox{\footnotesize orig}}}

\newcommand{\beq}{\begin{equation}}
\newcommand{\eeq}{\end{equation}}

\newtheorem{theorem}{Theorem}
\newtheorem{definition}{Definition}

\title{Non-negative Sparse Coding}
\author{Patrik O.\ Hoyer\\
        Neural Networks Research Centre\\
        Helsinki University of Technology\\
	P.O. Box 9800, FIN-02015 HUT, Finland\\
        patrik.hoyer@hut.fi}

\begin{document}
\maketitle

\begin{abstract}
Non-negative sparse coding is a method for decomposing multivariate
data into non-negative sparse components. In this paper we briefly
describe the motivation behind this type of data representation
and its relation to standard sparse coding and non-negative 
matrix factorization. We then give a simple yet efficient 
multiplicative algorithm for finding the optimal values of the
hidden components. In addition, we show how the basis vectors can
be learned from the observed data. Simulations demonstrate the 
effectiveness of the proposed method.
\end{abstract}

\section{Introduction}

Linear data representations are widely used in signal processing and
data analysis. A traditional method of choice for signal 
representation is of course Fourier analysis, but also wavelet 
representations are increasingly being used in a variety of 
applications. Both of these methods have strong mathematical 
foundations and fast implementations, but they share the important 
drawback that they are not adapted to the particular data being 
analyzed.

Data-adaptive representations, on the other hand, are representations
that are tailored to the statistics of the data. Such representations 
are learned directly from the observed data by optimizing some measure 
that quantifies the desired properties of the representation. This 
class of methods include principal component analysis (PCA), 
independent component analysis (ICA), sparse coding, and non-negative
matrix factorization (NMF). Some of these methods have their roots
in neural computation, but have since been shown to be widely
applicable for signal analysis.

In this paper we propose to combine sparse coding and non-negative 
matrix factorization into \emph{non-negative sparse coding} (NNSC). Again,
the motivation comes partly from modeling neural information processing.
We believe that, as with previous methods, this technique will be found
useful in a more general signal processing framework.

\section{Non-negative sparse coding}

Assume that we observe data in the form of a large number of 
i.i.d.\ random vectors $\x_n$, where $n$ is the sample index. 
Arranging these into the columns of a matrix $\X$, then linear 
decompositions describe this data as
$\X \approx \A\bS$. The matrix $\A$ is called the \emph{mixing matrix},
and contains as its columns the \emph{basis vectors} (features) of the
decomposition. The rows of $\bS$ contain the corresponding 
\emph{hidden components} that give the contribution of each basis vector 
in the input vectors. Although some decompositions provide an exact
reconstruction of the data (i.e. $\X = \A\bS$) the ones that
we shall consider here are approximative in nature.

In linear sparse coding \cite{Harpur96,Olshausen96b}, the goal is 
to find a decomposition in which the hidden components are \emph{sparse},
meaning that they have probability densities which are highly peaked at zero 
and have heavy tails. This basically means that any given input vector 
can be well represented using only a few significantly non-zero hidden 
coefficients. Combining the goal of small reconstruction error
with that of sparseness, one can arrive at the following objective 
function to be minimized \cite{Harpur96,Olshausen96b}:
\begin{equation} \label{eq:sc}
C(\A,\bS) = \frac{1}{2}\|\X - \A\bS\|^2 + \lambda\sum_{ij} f(S_{ij}),
\end{equation}
where the squared matrix norm is simply the summed 
squared value of the elements, i.e. $\|\X-\A\bS\|^2 = 
\sum_{ij}[\X_{ij}-(\A\bS)_{ij}]^2$.
The tradeoff between sparseness and accurate reconstruction is controlled
by the parameter $\lambda$, whereas the form of $f$ defines how sparseness
is measured. To achieve a sparse code, the form of $f$ must be chosen 
correctly: A typical choice is $f(s) = |s|$, although often similar
functions that exhibit smoother behaviour at zero are chosen for 
numerical stability.

There is one important problem with this objective: As $f$ typically
is a strictly increasing function of the absolute value of its argument,
the objective can always be decreased by simply scaling up $\A$ and 
correspondingly scaling down $\bS$. The consequences of this 
is that optimization of (\ref{eq:sc}) with respect to both $\A$ and $\bS$ 
leads to the elements of 
$\A$ growing (in absolute value) without bounds whereas $\bS$ tends
to zero. More importantly, the solution found does not depend
on the second term of the objective as it can always be eliminated
by this scaling trick. In other words, some constraint on the scales
of $\A$ or $\bS$ is needed. Olshausen and Field \cite{Olshausen96b}
used an adaptive method to ensure that the hidden components had unit 
variance (effectively fixing the norm of the rows of $\bS$), whereas
Harpur \cite{HarpurPhD} fixed the norms of the columns of $\A$.

With either of the above scale constraints the objective (\ref{eq:sc}) 
is well-behaved and its minimization can produce useful decompositions
of many types of data. For example, it was shown in \cite{Olshausen96b}
that applying this method to image data yielded features closely 
resembling simple-cell receptive fields in the mammalian primary 
visual cortex. The learned decomposition is also similar to
wavelet decompositions, implying that it could be useful in applications
where wavelets have been successfully applied. 

In standard sparse coding, described above, the data is described as a
combination of elementary features involving both additive and
subtractive interactions. The fact that features can `cancel each
other out' using subtraction is contrary to the intuitive notion of
combining parts to form a whole \cite{LeeDD99}. Thus, Lee and Seung
\cite{LeeDD99,LeeDD01} have recently forcefully argued for
non-negative representations \cite{Paatero94}. Other arguments for
non-negative representations come from biological modeling
\cite{Hoyer03CNS,Hoyer02VR,LeeDD99}, where such constraints 
are related to the non-negativity of neural firing rates.
These non-negative representations assume that the input data
$\X$, the basis $\A$, and the hidden components $\bS$ are all non-negative.

Non-negative matrix factorization\footnote{Note that error measures 
other than the summed squared error were also considered 
in \cite{LeeDD99,LeeDD01}.} (NMF) can be performed by the minimization
of the following objective function \cite{LeeDD01,Paatero94}:
\begin{equation}
C(\A,\bS) = \frac{1}{2}\|\X - \A\bS\|^2
\end{equation}
with the non-negativity constraints 
$\forall ij: \; A_{ij}\geq 0, \; S_{ij}\geq 0$.
This objective requires no constraints on the scales of $\A$ or $\bS$. 

In \cite{LeeDD99}, the authors showed how non-negative matrix 
factorization applied to face images yielded features that corresponded
to intuitive notions of face parts: lips, nose, eyes, etc. This was
contrasted with the holistic representations learned by PCA and
vector quantization. 

We suggest that both the non-negativity constraints and the sparseness
goal are important for learning parts-based representations. Thus,
we propose to combine these two methods into non-negative sparse coding:
\begin{definition}
Non-negative sparse coding (NNSC) of a non-negative data matrix $\X$ 
(i.e.\ $\forall ij: \; X_{ij}\geq 0$) is given by the minimization of
\begin{equation} \label{eq:nnsc}
C(\A,\bS) = \frac{1}{2}\|\X - \A\bS\|^2 + \lambda\sum_{ij} S_{ij}
\end{equation}
subject to the constraints $\forall ij: \; A_{ij}\geq 0, \; S_{ij}\geq 0$ and
$\forall i: \|\ai\| = 1$, where $\ai$ denotes the $i$:th column of $\A$.
It is also assumed that the constant $\lambda\geq 0$.
\end{definition}
Notice that we have here chosen to measure sparseness by a linear 
activation penalty (i.e. $f(s) = s$). 
This particular choice is primarily motivated by the fact that this 
makes the objective function quadratic in $\bS$. This is useful
in the development and convergence proof of an efficient algorithm
for optimizing the hidden components $\bS$.

\section{Estimating the hidden components}

We will first consider optimizing $\bS$, for a given basis $\A$. As
the objective (\ref{eq:nnsc}) is quadratic with respect to $\bS$, and the 
set of allowed $\bS$ (i.e. the set where $S_{ij}\geq 0$) is convex, 
we are guaranteed that no suboptimal local minima exist. The global
minimum can be found using, for example, quadratic programming
or gradient descent. Gradient descent is quite simple to 
implement, but convergence can be slow. On the other hand, 
quadratic programming is much more complicated to implement. 
To address these concerns, we have developed a multiplicative 
algorithm based on the one introduced in \cite{LeeDD01} that is
extremely simple to implement and nonetheless seems to be quite
efficient. This is given by iterating the following update rule:

\begin{theorem} \label{theorem:multupdate}
The objective (\ref{eq:nnsc}) is nonincreasing under the update rule:
\begin{equation} \label{eq:updateS}
\bS^{t+1} = \bS^{t} \hspace{1mm}.\hspace{-1mm}* (\A^T\X) \hspace{1mm}./\hspace{1mm} (\A^T\A\bS^{t} + \lambda)
\end{equation}
where $.*$ and $./$ denote elementwise multiplication and 
division (respectively), and the addition of the scalar $\lambda$ is done 
to every element of the matrix $\A^T\A\bS^{t}$.
\end{theorem}
This is proven in the Appendix. As each element of $\bS$ is updated
by simply multiplying with some non-negative factor, it is guaranteed
that the elements of $\bS$ stay non-negative under this update rule.
As long as the initial values of $\bS$ are all chosen strictly positive,
iteration of this update rule is in practice guaranteed to reach the
global minimum to any required precision.

\section{Learning the basis}

In this section we consider optimizing the objective (\ref{eq:nnsc})
with respect to both the basis $\A$ and the hidden components $\bS$, under
the stated constraints. First, we consider the optimization of 
$\A$ only, holding $\bS$ fixed.

Minimizing (\ref{eq:nnsc}) with respect to $\A$ 
\emph{under the non-negativity constraint only} could be done exactly 
as in \cite{LeeDD01}, with a simple multiplicative update rule. 
However, the constraint of unit-norm columns of $\A$ complicates
things. We have not found any similarly efficient update rule that would
be guaranteed to decrease the objective while obeying the 
required constraint. Thus, we here resort to projected gradient
descent. Each step is composed of three parts:
\begin{enumerate}
\item $\A' = \A^t - \mu (\A^t\bS-\X)\bS^T$
\item Any negative values in $\A'$ are set to zero
\item Rescale each column of $\A'$ to unit norm, and then set $\A^{t+1} = \A'$.
\end{enumerate}
This combined step consists of a gradient descent step (Step 1) followed by
projection onto the closest point satisfying both the non-negativity and 
the unit-norm constraints (Steps 2 and 3). This projected gradient step 
is guaranteed to decrease the objective if the stepsize $\mu>0$ is 
small enough and we are not already at a local minimum. (In this case 
there is no guarantee of reaching the \emph{global} minimum, due to the 
non-convex constraints.)

In the previous section, we gave an update step for $\bS$, holding $\A$
fixed. Above, we showed how to update $\A$, holding $\bS$ fixed. To
optimize the objective with respect to both, we can of course 
take turns updating $\A$ and $\bS$. This yields the following
algorithm:\\[2mm]
\centerline{
\fbox{
\begin{minipage}{0.90\textwidth}
\vspace{2mm}
{\bf Algorithm for NNSC}
\begin{enumerate}
\item Initialize $\A^0$ and $\bS^0$ to random \emph{strictly positive} 
matrices of the appropriate dimensions, and rescale each column of $\A^0$ 
to unit norm. Set $t=0$.
\item Iterate until convergence:
\begin{enumerate}
\item $\A' = \A^t - \mu (\A^t\bS^{t}-\X)(\bS^t)^T$
\item Any negative values in $\A'$ are set to zero
\item Rescale each column of $\A'$ to unit norm, and then set $\A^{t+1} = \A'$.
\item $\bS^{t+1} = \bS^{t} \hspace{1mm}.\hspace{-1mm}* ((\A^{t+1})^T\X) \hspace{1mm}./\hspace{1mm} ((\A^{t+1})^T(\A^{t+1})\bS^{t} + \lambda)$
\item Increment $t$.
\end{enumerate}
\end{enumerate}
\vspace{2mm}
\end{minipage}
}
}

\section{Experiments}

To demonstrate how sparseness can be essential for learning a
parts-based non-negative representation, we performed a simple
simulation where the generating features were known. The interested
reader can find the code to perform these experiments (as well
as the experiments reported in \cite{Hoyer03CNS}) on the web at:\\
\centerline{{\ttfamily http://www.cis.hut.fi/phoyer/code/}}\\

\begin{figure}
\hspace{1.2mm}
\large \bf \sffamily{Features}
\hspace{7mm}
\large \bf \sffamily{Data}
\hspace{10mm}
\large \bf \sffamily{NNSC}
\hspace{7mm}
\large \bf \sffamily{NMF (6)}
\hspace{2mm}
\large \bf \sffamily{NMF (10)} \\[1.3mm]
\resizebox{22mm}{!}{
\includegraphics{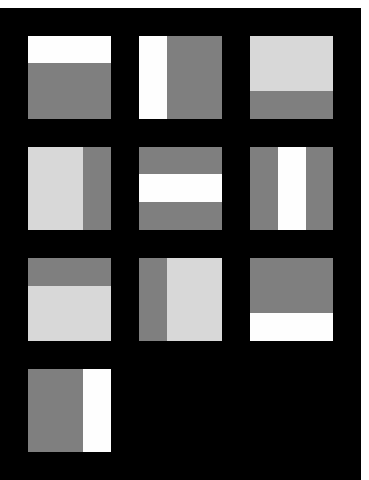}}
\resizebox{22mm}{!}{
\includegraphics{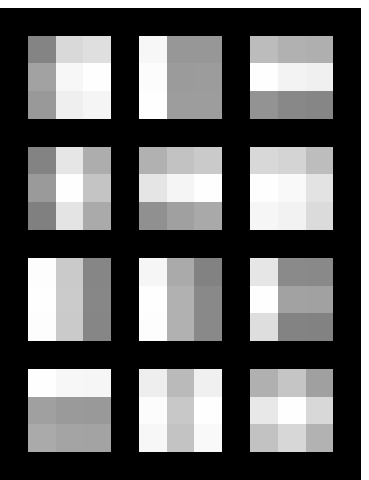}}
\resizebox{22mm}{!}{
\includegraphics{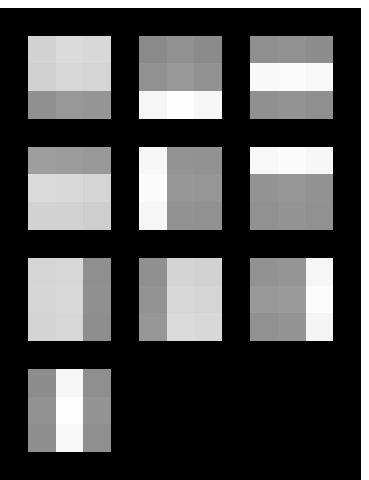}}
\resizebox{22mm}{!}{
\includegraphics{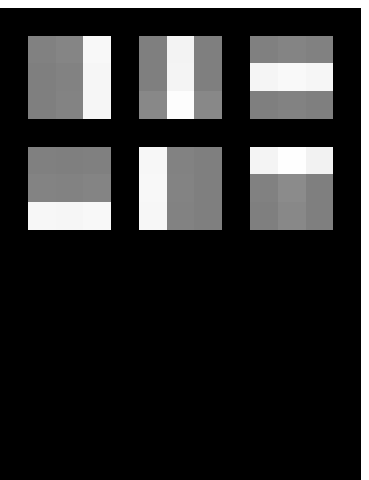}}
\resizebox{22mm}{!}{
\includegraphics{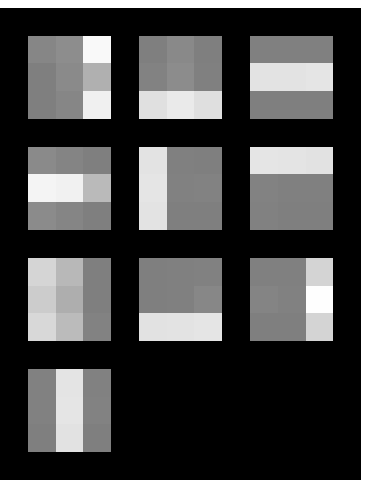}}
\caption{Experiments on bars data. {\mdseries  Features:} The $10$ original 
features that were used to construct the dataset. {\mdseries Data:} A random 
sample of 12 data vectors. These constitute superpositions of the 
original features. {\mdseries NNSC:} Features learned by NNSC, with 
dimensionality of the hidden representation equal to 10, starting from 
random initial values. {\mdseries NMF (6):} Features learned by NMF, 
with dimensionality 6. {\mdseries NMF (10):} Features learned by NMF, 
with dimensionality 10. See main text for discussion.\vspace{2mm}
\label{fig:experiments}}
\end{figure}

In our simulations, the data vectors were $3 \times 3$ -pixel images with
non-negative pixel values. We manually constructed $10$ original
features: the six possible horizontal and vertical bars, and the four
possible horizontal and vertical double bars. Each feature was
normalized to unit norm, and entered as a column in the matrix
$\Aorig$. The features are shown in the leftmost panel of 
Figure~\ref{fig:experiments}. We then generated random sparse non-negative 
data $\Sorig$, and obtained the data vectors as $\X = \Aorig\Sorig$. 
A random sample of $12$ such data vectors are also shown 
in Figure~\ref{fig:experiments}.

We ran NNSC and NMF on this data $\X$. With $10$ hidden components 
(rows of $\bS$),
NNSC can correctly identify all the features in the dataset. This result
is shown in Figure~\ref{fig:experiments} under {\bf \sffamily NNSC}. 
However, NMF cannot find all the features with any hidden dimensionality. 
With $6$ components, NMF finds all the single bar features. With 
a dimensionality of $10$, not even all of the single bars are correctly 
estimated. These results are illustrated in the two rightmost panels
of Figure~\ref{fig:experiments}.

It is not difficult to understand why NMF cannot learn all the
features.  The data $\X$ can be perfectly described as an additive
combination of the six single bars (because all double bars can be
described as two single bars). Thus, NMF essentially achieves the
optimum (zero reconstruction error) already with $6$ features, and
there is no way in which an \emph{overcomplete} representation could
improve that. However, when sparseness is considered as in NNSC, it is 
clear that it is useful to have double bar features because these 
allow a sparser description of such data patterns.

In addition to these simulations, we have performed experiments with
natural image data, reported elsewhere \cite{Hoyer03CNS,Hoyer02VR}.
These confirm our belief that sparseness is important when learning
non-negative representations from data.

\section{Relation to other work}

In addition to the tight connection to linear sparse coding 
\cite{Harpur96,Olshausen96b} and non-negative matrix factorization 
\cite{LeeDD99,LeeDD01,Paatero94}, this method is intimately related
to independent component analysis \cite{Hyva01book}. In fact, when
the fixed-norm constraint is placed on the rows of $\bS$ instead
of the columns of $\A$, the objective (\ref{eq:nnsc}) could be
directly interpreted as the negative joint log-posterior of the 
basis vectors and components, given the data $\X$, in the noisy 
ICA model \cite{Hoyer02VR}. This connection is valid when the independent
components are assumed to have exponential distributions, and of course the 
basis vectors are assumed to be non-negative as well.

Other researchers have also recently considered the constraint of 
non-negativity in the context of ICA. In particular, Plumbley 
\cite{Plumbley02SPL} has considered estimation of the noiseless
ICA model (with equal dimensionality of components and observations) 
in the case of non-negative components. On the other hand, 
Parra et al.\ \cite{Parra00} considered estimation of the ICA
model where the basis (but not the components) was constrained to be 
non-negative. The main novelty of the present work is the
application of the non-negativity constraints in the sparse coding
framework, and the simple yet efficient algorithm developed to estimate
the components.

\section{Conclusions}

In this paper, we have defined non-negative sparse coding as a combination
of sparse coding with the constraints of non-negative
matrix factorization. Although this is essentially a special case
of the general sparse coding framework, we believe that the proposed
constraints can be important for learning parts-based representations
from non-negative data. In addition, the constraints allow a very
simple yet efficient algorithm for estimating the hidden components.

\section{Appendix}

To prove Theorem~\ref{theorem:multupdate}, first note that 
the objective (\ref{eq:nnsc}) is separable in the
columns of $\bS$ so that each column can be optimized without
considering the others. We may thus consider the problem for the case
of a single column, denoted $\s$. The corresponding column of $\X$ is
denoted $\x$, giving the objective
\begin{equation}
F(\s) = \frac{1}{2}\|\x - \A\s\|^2 + \lambda\sum_i s_i.
\end{equation}

The proof will follow closely the proof given 
in \cite{LeeDD01} for the case $\lambda=0$. 
(Note that in \cite{LeeDD01}, the notation $v=\x$, $W=\A$ and $h=\s$ 
was used.)
We define an auxiliary function $G(\s,\s^t)$
with the properties that $G(\s,\s) = F(\s)$ and $G(\s,\s^t)\geq F(\s)$.
We will then show that the multiplicative update rule corresponds to
setting, at each iteration, the new state vector to the values
that minimize the auxiliary function:
\beq
\s^{t+1} = \arg\min_{\s} G(\s,\s^t).
\eeq
This is guaranteed not to increase the objective function $F$, as
\beq
F(\s^{t+1}) \leq G(\s^{t+1},\s^t) \leq G(\s^t,\s^t) = F(\s^t).
\eeq 

Following \cite{LeeDD01}, we define the function $G$ as
\beq 
\label{eq:auxdef}
G(\s,\s^t) = F(\s^t) + (\s-\s^t)^T\nabla F(\s^t) + 
\frac{1}{2}(\s-\s^t)^T\K(\s^t)(\s-\s^t)
\eeq
where the diagonal matrix $\K(\s^t)$ is defined as
\beq
K_{ab}(\st) = \delta_{ab} \frac{(\A^T\A\s^t)_a + \lambda}{\s^t_a}.
\eeq
It is important to note that the elements of our choice for $\K$ are 
always greather than or equal to those of the $\K$ used in \cite{LeeDD01}, 
which is the case where $\lambda=0$. It is obvious that 
$G(\s,\s) = F(\s)$. Writing out
\beq
F(\s) = F(\s^t) + (\s-\s^t)^T\nabla F(\s^t) + 
\frac{1}{2}(\s-\s^t)^T(\A^T\A)(\s-\s^t),
\eeq
we see that the second property, $G(\s,\s')\geq F(\s)$, 
is satisfied if
\beq
0 \leq (\s-\s^t)^T[\K(\s^t)-\A^T\A](\s-\s^t).
\eeq
Lee and Seung proved this positive semidefiniteness
for the case of $\lambda=0$ \cite{LeeDD01}. In our case, with $\lambda>0$,
the matrix whose positive semidefiniteness is to be proved is the same 
except that a strictly non-negative diagonal matrix has been added
(see the above comment on the choice of $\K$). As a non-negative 
diagonal matrix is positive semidefinite, and the sum
of two positive semidefinite matrices is also positive semidefinite,
the $\lambda=0$ proof in \cite{LeeDD01} also holds when $\lambda>0$.

It remains to be shown that the update rule in (\ref{eq:updateS})
selects the minimum of $G$. This minimum is easily found by taking 
the gradient and equating it to zero:
\beq
\nabla_{\s} G(\s,\s^t) = \A^T(\A\s^t-\x) + \lambda\cvec 
+ \K(\s^t)(\s-\s^t) = 0,
\eeq
where $\cvec$ is a vector with all ones. Solving for $\s$, this gives
\begin{eqnarray}
\s & = & \s^t - \K^{-1}(\s^t)(\A^T\A\s^t - \A^T\x + \lambda\cvec) \\
   & = & \s^t - (\s^t ./ (\A^T\A\s^t + \lambda\cvec)) 
         .\hspace{-1mm}* (\A^T\A\s^t - \A^T\x + \lambda\cvec) \\
   & = & \s^t .\hspace{-1mm}* (\A^T\x) ./ (\A^T\A\s^t + \lambda\cvec))
\end{eqnarray}
which is the desired update rule (\ref{eq:updateS}). This completes
the proof.

\bibliography{/home/info/phoyer/research/bib/collection,/home/info/phoyer/research/bib/others,/home/info/phoyer/research/bib/personal}

\begin{thebibliography}{10}
\newcommand{\enquote}[1]{``#1''}

\bibitem{HarpurPhD}
G.~F. Harpur, \textbf{Low Entropy Coding with Unsupervised Neural Networks},
  Ph.D.\ thesis, University of Cambridge, 1997.

\bibitem{Harpur96}
G.~F. Harpur and R.~W. Prager, \enquote{Development of low entropy coding in a
  recurrent network,} \textbf{Network: Computation in Neural Systems}, vol.~7,
  pp.~277--284, 1996.

\bibitem{Hoyer03CNS}
P.~O. Hoyer, \enquote{Modeling receptive fields with non-negative sparse
  coding,} Submitted. Available online, see {\ttfamily
  http://www.cis.hut.fi/phoyer/papers/}.

\bibitem{Hoyer02VR}
P.~O. Hoyer and A.~Hyv\"{a}rinen, \enquote{A multi-layer sparse coding network
  learns contour coding from natural images,} \textbf{Vision Research}, In
  press. Available online, see {\ttfamily
  http://www.cis.hut.fi/phoyer/papers/}.

\bibitem{Hyva01book}
A.~Hyv\"arinen, J.~Karhunen and E.~Oja, \textbf{Independent Component
  Analysis}, Wiley Interscience, 2001.

\bibitem{LeeDD99}
D.~D. Lee and H.~S. Seung, \enquote{Learning the parts of objects by
  non-negative matrix factorization,} \textbf{Nature}, vol.~401, no. 6755,
  pp.~788--791, 1999.

\bibitem{LeeDD01}
D.~D. Lee and H.~S. Seung, \enquote{Algorithms for non-negative matrix
  factorization,} in \textbf{Advances in Neural Information Processing 13
  (Proc.~NIPS*2000)}, MIT Press, 2001.

\bibitem{Olshausen96b}
B.~A. Olshausen and D.~J. Field, \enquote{Emergence of simple-cell receptive
  field properties by learning a sparse code for natural images,}
  \textbf{Nature}, vol.~381, pp.~607--609, 1996.

\bibitem{Paatero94}
P.~Paatero and U.~Tapper, \enquote{Positive Matrix Factorization: A
  Non-negative Factor Model with Optimal Utilization of Error Estimates of Data
  Values,} \textbf{Environmetrics}, vol.~5, pp.~111--126, 1994.

\bibitem{Parra00}
L.~Parra, C.~Spence, P.~Sajda, A.~Ziehe and K.-R. M\"{u}ller, \enquote{Unmixing
  Hyperspectral Data,} in \textbf{Advances in Neural Information Processing 12
  (Proc.~NIPS*99)}, MIT Press, pp. 942--948, 2000.

\bibitem{Plumbley02SPL}
M.~Plumbley, \enquote{Conditions for non-negative independent component
  analysis,} \textbf{IEEE Signal Processing Letters}, Submitted.

\end{thebibliography}
\bibliographystyle{nnsp}

\section{Acknowledgements}

I wish to acknowledge Aapo Hyv\"{a}rinen for useful discussions and 
helpful comments on an earlier version of the manuscript.

\end{document}